\documentclass[11pt]{article}
\usepackage{eacl2006}
\usepackage{times}
\usepackage{amssymb}
\usepackage{latexsym}
\usepackage{graphicx}
\title{Lexical Adaptation of Link Grammar to the Biomedical Sublanguage:\\a Comparative Evaluation of Three Approaches}

\author{Sampo Pyysalo \and  Tapio Salakoski\\
Turku Centre for Computer Science (TUCS)\\
and University of Turku \\
Lemmink\"{a}isenkatu 14 A,\\ FIN-20520 Turku, Finland  
 \And
 Sophie Aubin \and  Adeline Nazarenko\\
 LIPN\\ Universit\'e Paris 13 \& CNRS UMR 7030\\
99, av. J.-B. Cl\'ement,\\ F-93430 Villetaneuse, France\\
}

\date{\today}

\begin{document}
\maketitle
\begin{abstract}
We study the adaptation of Link Grammar Parser to the biomedical
sublanguage with a focus on domain 
terms not found in a general
parser lexicon.  Using two biomedical corpora, we implement and evaluate
three approaches to addressing unknown words: automatic lexicon expansion,
the use of morphological clues, and disambiguation using a
part-of-speech tagger.  We evaluate each approach separately for its
effect on parsing performance and consider combinations of these
approaches. In addition to a 45\% increase in parsing efficiency, we
find that the best approach, incorporating information from a domain 
part-of-speech tagger, offers a statistically significant 10\%
relative decrease in error. 
The adapted parser is available 
under an open-source license at 
\texttt{http://www.it.utu.fi/biolg}.
\end{abstract}


\section{Introduction}

In applying general parsers to specific domains, adaptation is
often necessary to achieve high parsing performance 
(see e.g.\ \cite{sekine97parsing}). Sublanguage is 
defined by Grishman~\shortcite{grishman01sublanguage}
as a specialized form of a natural language that is used within a
particular domain or subject matter. It is characterized by 
specialized vocabulary, semantic relationships, and in many cases
 syntax. 

In this paper, we study lexical adaptation, that is, adaptation
addressing the specialized vocabulary.  This is an important part of
the process of customizing a general parser to a sublanguage.
Among other issues, the unknown word rate increases dramatically when
moving from general language to increasingly technical domains such
as that of biomedicine \cite{lease05parsing}.
This can lead to increased ambiguity, reduced parsing performance, and
errors in establishing the correct relationships between words for
semantic mining \cite{pyysalo2006evaluation}.

Until recently,
Information Extraction (IE) systems for mining semantic
relationships from texts of
technical sublanguages
avoided full syntactic parsing. The quality of parsing has a 
well-established effect on the performance of IE systems, and
the accuracy of general parsers in technical domains is 
comparatively low. Additionally, many domain-specific parsers
lack portability to a new domain.
Finally, the time required for full parsing is also a
problem for IE systems. But the biomedical IE community now faces
limitations in pattern-matching \cite{blaschke99extraction} and
shallow parsing \cite{pustejovsky02parsing} methods that are
inefficient in the processing of long distance dependencies 
and complex
sentences.
Recent advances in parsing techniques have further created an
increased interest in the adaptation of full parsers.

Here, we consider the lexical adaptation of a full parser, the Link
Grammar Parser\footnote{http://www.link.cs.cmu.edu/link/} (LGP) of
Sleator of Temperley~\shortcite{sleator91parsing}.  The choice of
parser addresses the recent interest in LGP in the biomedical IE
community
\cite{ding03extracting,szolovits03adding,ahmed05IntEx,alphonse2004caderige}.
Our evaluation is performed using two corpora of sentences from
Medline abstracts 
with a focus on protein-protein interactions,
the identification of which is the key aim of most biomedical IE
systems.

Recently, two approaches addressing unknown words in applying LGP 
to the biomedical domain have been proposed.
Szolovits~\shortcite{szolovits03adding} introduced a method for
heuristically mapping terminology between lexicons and applied this
mapping to augment the LGP dictionary with terms from the UMLS
Specialist Lexicon\footnote{http://specialist.nlm.nih.gov/}. 
Based on an analysis of a domain corpus, two of the authors have
proposed an extension of the morpho-guessing system of LGP for 
disambiguating domain terms based on their suffixes
\cite{aubin05adaptation}.
The effect of the proposed extensions on parsing performance against
an annotated reference corpus was not evaluated in these two studies.

Here we analyze the effect of these lexical extensions
using an annotated biomedical corpus. We further propose, implement
and evaluate in detail a third approach to resolving unknown words in
LGP using information from a part-of-speech (POS) tagger.

\section{Link Grammar Parsing}

The link grammar formalism is closely
related to dependency formalism
. It is based on the notion of typed
\emph{links} connecting words.
The result of parsing is one or more ordered parses, termed \emph{linkages}.
A linkage 
consists of a set of links
connecting the words of a sentence so that links do not cross, no two
links connect the same two words, and the types of the links satisfy
the \emph{linking requirements} given to each word in the lexicon.
An example linkage is given below.\\
  \includegraphics[scale=0.9]{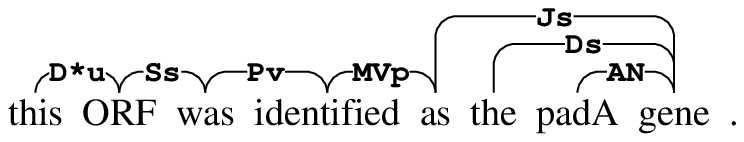} \\
Since the link grammar is rule-based and the parser makes no use of statistical
methods, LGP is a good candidate for
adaptation to new domains where annotated corpus data is rarely 
available. 
The lexical adaptation approaches we evaluate further
require only a light linguistic analysis of 
domain language.

LGP has three different methods applied in a cascade 
to handle
vocabulary: dictionary lookup, morpho-guessing and unknown word
guessing. 
The LGP dictionary 
enumerates all words, including inflected forms, and
grammar rules are encoded through the linking requirements associated
with the words.
Some unknown words are assigned linking requirements based on their
morphological features, such as the suffix \emph{-ly} for adverbs.
This system is termed \emph{morpho-guessing} (MG). 
Finally, words that are neither found in the parser dictionary nor 
recognized by its morpho-guessing rules are 
assigned all possible combinations of the generic verb, noun
and adjective linking requirements. 
This general approach is, in
principle, always capable of generating the correct combination of
linking requirements for unknown words. However, with an increasing 
number of unknown words in a sentence, the approach leads to a
combinatorial explosion in the number of possible linkages and a
rapid increase in parsing time and decrease in parsing performance.  
The parser is also time-limited: when a sentence cannot be parsed
within a user-specified time limit, LGP attempts parses using
more efficient, but restricted settings, leading to reduced
parse quality.


When parsing 
sublanguages
that contain many
words that are not in the lexicon, it is therefore beneficial to attempt
to resolve unknown words to reduce ambiguity in parsing.

\section{Lexical adaptations}

We evaluate three approaches to lexical adaptation: lexicon extension,
morphological clues, and POS tagging.  The approaches primarily
involve open-class words and use linking requirements from the
original LGP.
Closed-class words, such as prepositions are considered 
domain-independent and expected to appear in the original lexicon,
and modification of the existing linking requirements (grammar adaptation)
is outside the scope of this study.
\begin{table*}[!htbp]
 \centering
  \begin{small}
  \begin{tabular}{l|l|l||l|l|l}
    Suffix & POS & examples & Suffix & POS & examples\\
   \hline
    \textit{-ase} & noun &  synthetase, kinase &
    \textit{-in} & noun & actin, kanamycin\\
    \textit{-ity} & noun & chronicity, hypochromicity &  
     \textit{-ion} & noun & septation, reguion  \\
    \textit{-on} & noun & replicon, intron & 
    \textit{-ol} & noun & glycosylphosphatidylinositol \\
    \textit{-ose} & noun & isomaltotetraose, isomaltotriose
    & \textit{-or} & noun & cofactor, repressor/activator \\
    \textit{-yl} & noun &  hydroxyethyl, hydroxymethyl
    & \textit{-ine} & noun & 5-(hydroxymethyl)-2'-deoxyuridine \\
    \textit{-ide} & noun & iodide, oligodeoxynucleotide 
    & \textit{-i} & noun & casei, lactococci, termini\\
    \textit{-ic} & adjective & glycolytic, ribonucleic, uronic
    & \textit{-al} & adjective & ribosomal, ribsosomal 
    \\
   \textit{-ive} & adjective & nonpermissive, thermosensitive 
   & \textit{-ar} & adjective & intermolecular, intramolecular\\
   \textit{-ble} & adjective & inducible, metastable 
   & \textit{-ous} & adjective & exogenous, heterologous \\
   \textit{-ae} & latin adj. & influenzae, tarentolae 
   & \textit{-us} & latin adj. & pentosaceus, luteus, carnosus \\
   \textit{-um} & latin adj. & japonicum, tabacum, xylinum 
   & \textit{-is} & latin adj. & brevis, israelensis 
   \\
   \textit{-fold} & adjective/adverb & 10-fold, 4.5-fold, five-fold \\
  \end{tabular}
  \caption{Biomedical suffixes involved  in the extension of the
    morpho-guessing rules}
  \label{tab:MGextension}
  \end{small}
\end{table*}

\subsection{Extension of the lexicon}

The extension of the lexicon with external domain-specific knowledge is
the most frequent approach to adaptation, provided that the resources are
available for the domain. This can be done either manually 
or with automatic mapping methods.  


Here, we evaluate the heuristic lexicon
mapping proposed by Szolovits~\shortcite{szolovits03adding}.
This mapping can be used to automatically add domain-specific terminology
from an external specialized lexicon to the lexicon of a parser.
Words are mapped from a source lexicon (e.g.\ the domain lexicon)
to a target lexicon (e.g.\ the parser lexicon) based on their lexical
descriptions. As these descriptions typically differ between lexicons,
they cannot be transferred directly from one lexicon
to another. 
Instead, the mapping operates with sets of words that have the
exact same lexical description in their respective lexicons.

To assign a lexical description to a word $w$ not in the target
lexicon, the mapping finds words that have the exact same lexical
description as $w$ in the source lexicon, and that further have a
description in the target lexicon. Overlap in sets having the same
descriptions is then used to select one of these target lexicon
descriptions to assign to $w$.

Szolovits applied the introduced mapping to extend the lexicon of LGP
with terms from the UMLS Specialist Lexicon and observed that the mapping
heuristic chose poor definitions for some smaller sets, for which the
definitions were manually modified.  The created UMLS dictionary
extension contains 121,120 words that do not appear in the original LGP 
dictionary.

Szolovits observed that many of the phrases included in the
extension ``bear no specific lexical information in Specialist that is
not obvious from their component words''. Additionally, phrases are
parsed using the LGP idiom system, which does not assign internal
structure to the phrases, complicating comparison against a reference
corpus.  For these reasons, we evaluate the \texttt{no-phrases}
version of the extension\footnote{
http://www.cdm.csail.mit.edu/projects/text/}. The effect of this
extension has also been considered by Pyysalo et~al.~\shortcite{pyysalo2006evaluation}.

\subsection{Morphological clues} 

Morphological clues can be exploited by LGP
to predict the morpho-syntactic classes (hence syntactic behaviour) 
of unknown words.
Specific domains are an interesting application for this type of
adaptation because a great part of technical lexicons presents regular
morphological features, which, according to
Mikheev~\shortcite{mikheev97guessing}, obey morphological regularities
of the general language. We observe that this assumption holds only
partially because of the presence of foreign words in specialized
texts and argue that a minimal morphological study of the corpus is
necessary. Such studies have been performed, on the biomedical domain
by Spyns~\shortcite{spyns94medical} and Aubin et~al.~\shortcite{aubin05adaptation}.

While many POS taggers employ morphological features to tag unknown
words, domain extension of a rule-based approach such as the LGP
morpho-guessing system 
can be preferable in lexical adaptation to
domains where resources such as tagged corpora 
are not available for training taggers. Further, the MG extension 
allow assigning specific rules at a greater granularity than POS 
tags.

We have implemented and evaluated the extension of the LGP morpho-guessing rules proposed by
Aubin et~al.~\shortcite{aubin05adaptation}. 
This extension of 23 new suffixes for the
biomedical domain is presented in Table~\ref{tab:MGextension}. 
Aubin et~al.~\shortcite{aubin05adaptation} further identified in the corpus a small number of
exceptions to these rules (\textit{"wherein"}, \textit{"kcal/mol"},
\textit{"ultrafine"}, etc.), which were
manually added to the dictionary.

 

\subsection{POS tagging} 
We finally propose to provide the parser with an input
sentence enriched with POS tags. In order to retain the
decision-making power of the parser and to avoid  inconsistencies
between tagged words and their entry in the parser
lexicon (see Grover et al.~\shortcite{grover05parsing}), we restrict the use of POS tags to unknown words only.


We modified LGP so that POS 
information can be passed to the parser by appending POS tags to
input words (e.g.\ \emph{actin/NN}). We further modified
the parser so that when an unknown word is given a POS tag, the
parser assigns linking requirements to the words based on a given
mapping from POS tags to LGP dictionary entries.
We defined such a mapping, presented in Table \ref{table:POSmap}, for 
Penn tagset POS categories corresponding to content words. FW (foreign words)
 and SYM (symbols) tags were not mapped due to their syntactic
heterogeneity. Existing LGP rules were used to define the behavior
of POS-mapped words, and the most generic applicable rule was chosen
in each case. For instance, words tagged "NN" map the rule for nouns 
that can be either mass or countable, so that there is no constraint on
determiners. 

\begin{table}
\centering
  \begin{small}
\begin{tabular}{l|l|l}
Tag  & Description & LGP rule \\
\hline
NN   & common noun, sing.  & words.n.4 \\
NNS  & common noun, pl.   & words.n.2.s \\
NNP  & proper noun, sing.   & \scriptsize{CAPITALIZED-WORDS} \\
NNPS & proper noun, pl.   & \scriptsize{PL-CAPITALIZED-WORDS} \\
JJ  & adjective, base & \scriptsize{UNKNOWN-WORD.a} \\
JJR  & adjective, comparative  & words.adj.2 \\
JJS  & adjective, superlative  &words.adj.3 \\
VB   & verb, base  & words.v.6.1 \\
VBD  & verb, past tense & words.v.6.3 \\
VBZ  & verb, present 3d pers. & \scriptsize{S-WORDS.v} \\
VBP  & verb, present non-3d & words.v.6.1 \\
VBG  & verb, gerund & \scriptsize{ING-WORDS} \\
VBN  & verb, past participle & \scriptsize{ED-WORDS}\\
CD   & number & \scriptsize{NUMBERS} \\
RB   & adverb, base & words.adv.1 \\
\end{tabular}
\caption{POS tags mapping to LGP rules}\label{table:POSmap}
\end{small}
\end{table} 


To evaluate the effect of using both a general and a domain tagger,
the experiments were made using two taggers: the Brill
tagger\footnote{http://research.microsoft.com/users/brill/} trained on
the Wall Street Journal (general language) and the
GENIA Tagger\footnote{http://www-tsujii.is.s.u-tokyo.ac.jp/GENIA/tagger/}
\cite{tsuruoka2005tagger} trained on the biomedical corpus
GENIA. A detailed evaluation and error analysis of GENIA Tagger
is given in \cite{tsuruoka2005tagger}, finding 98\% accuracy on
two biomedical corpora. On this basis, 
we estimate the tagging accuracy at 81\% for the Brill tagger and 97\%
for GENIA Tagger.  This estimate was performed by manually checking
tagging divergences between the two taggers on one of our corpora.
\section{Evaluation protocol}

\subsection{Corpora}

Two corpora are used for the present evaluation: ``interaction'' and
``transcript'', both built in the context of IE in biomedical texts. 
Both corpora were tokenized and cleared of bibliographic
references in a preprocessing step.

Interaction contains 542 sentences (16,874 tokens) annotated for
dependencies using the Link Grammar annotation scheme.  600 sentences
were initially selected randomly from
Pubmed\footnote{http://www.pubmed.com/} with the condition that they
contain at least two proteins for which a known interaction was
entered into the DIP database\footnote{http://dip.doe-mbi.ucla.edu/}.
58 sentences consisting only of a nominal phrase were then excluded as
LGP does not, by design, parse them\footnote{This limitation could be overcome
by modification of the grammar, but here we decided to avoid grammar
adaptation and evaluate the parser with respect to its intended coverage.}.
Each sentence was separately annotated by two annotators, and differences
were resolved by discussion.  Links to punctuation were excluded, and
link types were not annotated.  A total of 14,242 links were annotated
in these sentences.




The transcript corpus is made of 16,989 sentences (438,390 tokens) consisting of
the result for the query \textit{``Bacillus subtilis transcription''}
on Pubmed. It was not annotated.


Both corpora are used to characterize the vocabulary coverage by the different methods
applied in LGP. The annotated interaction corpus is also
used as the reference corpus for the
evaluation of parsing performance.

\begin{figure*}
\includegraphics[scale=0.65]{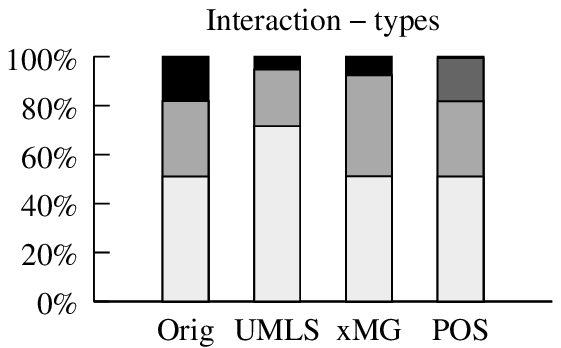} 
\includegraphics[scale=0.65]{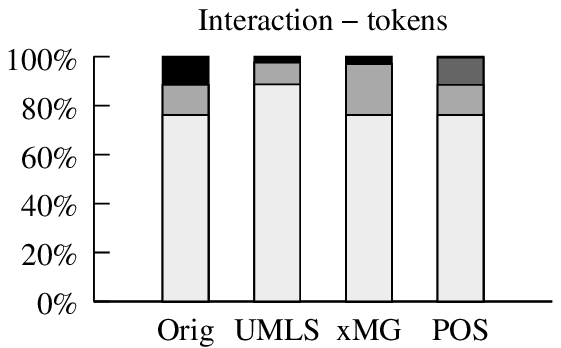}
\includegraphics[scale=0.65]{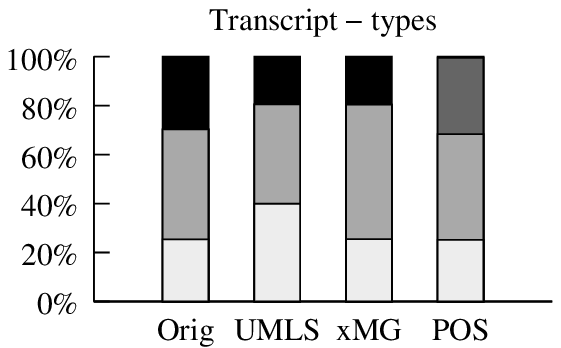} 
\includegraphics[scale=0.65]{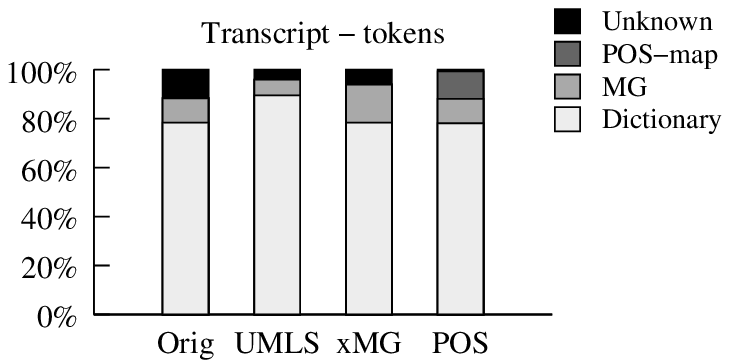}
\caption{Vocabulary handling in the interaction and transcript corpora:
the fraction of words and types covered by each method in the original
LGP and the three adaptations.
Coverage for the POS adaptation is shown only for GENIA Tagger as the coverage of the Brill
tagger was essentially identical.}
\label{fig:lex_both}
\end{figure*}

\subsection{Evaluation criteria}
We first evaluate 
\emph{vocabulary coverage} in the 
original and extended versions of LGP.  
We present the contribution of each method (dictionary, morpho-guessing, POS-mapping and unknown 
words) implemented in LGP 
to handle vocabulary. 
Results are given 
separately for 
types (i.e.\ distinct forms) and tokens (i.e.\ occurrences) in the corpus.

We assess the \emph{ambiguity of the parsing} process with two 
criteria: parsing time and linkage numbers.
Parsing time is immediately relevant to applications of the parser to
systems where large
corpora must be parsed. Linkage numbers 
are a
more direct measure of the ambiguity of parsing a sentence.  For each
sentence, the parser enumerates the total number of linkages allowed
by the grammar. By taking the ratio of the number of linkages allowed
by two versions of the parser, we can estimate the relative increase
or decrease in ambiguity. 
We report the per-sentence averages of both parsing time and
linkage number ratios.

To determine the \emph{parsing performance} of the extensions of LGP, 
we used each of the extensions to parse the interaction corpus
sentences and compared the produced linkages against the
reference corpus. 
For each sentence, we determine 
the recall, i.e.
the fraction of links in the 
reference corpus 
that were present in parses returned by 
LGP\footnote{Note that for connected, acyclic dependency graphs,
precision equals recall: for each missing link, there is exactly one
extra link. While there are some exceptions to connectedness and
acyclicity in both LGP linkages and the annotation, we believe
recall can be used as a fair estimate of overall performance.}.
We report average recall for both the \emph{first linkages} as ordered by
the LGP heuristics and, to separate the effect of the heuristics from
parser performance, also the \emph{best linkages}, that is,
the linkages with the most annotated links recovered.
We further separately evaluate overall performance and performance
for the subset of sentences where no timeouts occurred in parsing.

Experiments were performed on a 2.8GHz Intel Xeon with
parameter values \texttt{timeout=60sec}, \texttt{limit=1000},
\texttt{islands-ok=true}.
Default values were used for other parameters.  The statistical
significance of differences between the original parser and each of
the modifications is assessed using the Wilcoxon signed-ranks test for
overall first linkage performance, using the Bonferroni correction for
multiple comparisons.

\section{Results}

In this section we present the evaluation results for the original
LGP (Orig), LGP with the UMLS dictionary extension (UMLS), LGP with
the morpho-guessing extension (xMG) and LGP with the POS extension,
evaluated with the two taggers, Brill and GENIA tagger (GT).

\subsection{Vocabulary coverage}

Figure \ref{fig:lex_both} shows 
the proportion of vocabulary covered by each method on 
the interaction and transcript corpora .


The comparison of the results on 
types and tokens
shows that the dictionary 
has a good recognition rate on frequent types for both the original and the UMLS versions. By contrast, the MG and POS-map methods contribute for the recognition of a great number of types (particularly in transcript) but few tokens. 
In addition, the discrepancy on types between the two corpora for the dictionary method in all versions reflects the increasing presence of low frequency non-canonical words with the growing size of the corpus. 
Interestingly, we find that the reduction in unknown words (black part
in the charts) due to the
UMLS and xMG extensions is roughly similar, despite the former containing
over 100,000 new words and the latter only 23 new rules.
The POS extension, as expected, reduces the part of unknown words to almost null. 

The remaining unknown words are of different nature for the extensions. Quite surprisingly, UMLS lacks a great number of species names (numerous in transcript) and frequent gene or protein names (e.g.\ \textit{lacZ}, 78 occurrences in transcript). In addition, the Specialist Lexicon version used here contains no complex terms which prevents from detecting words like \textit{vitro} and \textit{vivo} used in the frequent terms \textit{in vitro} and \textit{in vivo}. The 
evaluated xMG extension cannot handle gene/protein names either, and
also misses frequent technical terms that have no specific
morphological features, such as \textit{sigma, mutant} and
\textit{plasmid}.


\begin{table*}[ht]
\centering
\begin{tabular}{l|r|rr|rr|rr|rr}
                       & Orig & UMLS & $\Delta$ & xMG & $\Delta$ & Brill & $\Delta$ & GT & $\Delta$ \\ \hline
All, first linkage & 74.2 & 75.4 & 4.7  & 76.0 & 7.0  & 75.4 & 4.7  & 76.8 & 10.1\\
All, best linkage & 82.7 & 83.5 & 4.6  & 84.5 & 10.4 & 83.7 & 5.8  & 85.3 & 15.0\\
NT, first linkage & 78.0 & 78.1 & 0.5  & 78.9 & 4.1  & 78.0 & 0.0  & 79.4 & 6.4 \\
NT, best linkage  & 87.4 & 86.9 & -4.0 & 88.0 & 4.8  & 86.7 & -5.6 & 88.3 & 7.1 \\
\hline
$p$            & N/A & \multicolumn{2}{r|}{$p\approx 0.06$} & \multicolumn{2}{r|}{$p<0.01$} & \multicolumn{2}{r|}{$p\approx 0.07$} & \multicolumn{2}{r}{$p<0.01$} \\
\end{tabular}
\caption{Performance. First linkage denotes the linkage ordered first
by the parser heuristics and best linkage the best performance
achieved by any linkage returned by the parser. Results marked NT are
for the subset of sentences where no timeouts occurred for any of the
modifications. $\Delta$ columns give relative decrease in error with
respect to the original LGP, and $p$ values are
for ``All, first linkage'' performance.
}\label{table:simple_performance}
\end{table*}

To assess lexicon coverage, we measured the
\emph{contribution}\footnote{proportion of types of the resource found
in the corpus} and the \emph{recognition}\footnote{proportion of types
of the corpus found in the resource} of the UMLS dictionary extension.
We find that while
the contribution of the UMLS dictionary extension is very low, with
0.54\% on interaction and 2.3\% on transcript, the recognition of the
dictionary method is augmented significantly by the UMLS extension
(51\% to 71\% for interaction and 25\% to 40\% for transcript). 
Nevertheless, as the size of the dictionary does not significantly
penalize the parsing time with LGP, even a generic resource that
contributes relatively little can be beneficial.



\subsection{Ambiguity}

The results of measuring the effect of the various extensions on ambiguity 
are given in Table~\ref{table:simple_ambiguity}.

\begin{table}[ht]
\centering
\begin{small}
\begin{tabular}{l|r|r|r|r|r}
Metric               & Orig  & UMLS  & xMG    & Brill  & GT    \\ \hline
Time                 & 15.4s & 9.9s  & 10.8s & 8.8s   & 8.6s  \\
Lkg.\ ratio          & 1     & 0.67  & 0.68  & 0.70   & 0.66  \\
\end{tabular}
\caption{Ambiguity. Time is average parsing time per sentence, linkage ratio
is average of per-sentence linkage number ratios.}\label{table:simple_ambiguity}
\end{small}
\end{table}

The reduction in the number of unknown words for the 
UMLS and xMG extensions
is coupled with a
roughly 30\% reduction in both parsing time and linkage numbers.  Although
the POS extension essentially eliminates unknown words, 
it only gives a decrease in parsing time and linkage numbers that roughly
mirrors the effect of the UMLS and xMG extensions.

None of the extensions achieves more than 35\% reduction in linkage
numbers or more than 45\% reduction in parsing time.
This may
reflect 
structural ambiguity in the language and suggest a limit on
how much ambiguity can be controlled 
through these lexical adaptation approaches.

\subsection{Performance}
The evaluation results are presented in Table \ref{table:simple_performance}. 
We find that in addition to increased efficiency, all of the
extensions offer an
increase in overall parsing performance compared to the original LGP
for both the first and best linkages. 
Remarkably, this increase occurs even with
the Brill tagger, which was trained on general English. 
In overall performance, the UMLS extension and the POS extension
with the Brill tagger are roughly equal. The xMG extension outperforms
both, and the POS extension with GENIA Tagger has the best performance
of all considered extensions.

\begin{table*}[t!h]
\centering
\begin{tabular}{l|r|rr|rr|rr|rr}
                    &       & UMLS  &          & xMG     &          & UMLS   &          & \\
Metric              & Orig  & \& xMG & $\Delta$ & \& POS & $\Delta$ & \& POS & $\Delta$ & All 3 & $\Delta$   \\ \hline
All, first linkage& 74.2 & 75.7 & 5.8 & 76.8  & 10.1 & 76.0  & 7.0 & 76.1 & 7.4 \\
All, best linkage & 82.7 & 83.7 & 5.8 & 85.3  & 15.0 & 84.2  & 8.7 & 84.2 & 8.7 \\
NT, first linkage & 78.0 & 78.4 & 1.8 & 79.3  & 5.9  & 78.6  & 2.7 & 78.7 & 3.2 \\
NT, best linkage  & 87.4 & 87.0 & -3.2 & 88.2 & 6.3  & 87.2  & -1.6& 87.1 & -2.4\\
\hline
$p$            & N/A & \multicolumn{2}{r|}{$p<0.05$} & \multicolumn{2}{r|}{$p<0.01$}  & \multicolumn{2}{r|}{$p<0.01$} & \multicolumn{2}{r}{$p<0.01$} \\
\end{tabular}
\caption{Performance for combinations of the extensions.}
\label{table:combination_performance}
\end{table*}

The positive effect of the extensions on parsing performance is linked
to the reduced number of timeouts that occurred when parsing.
Effects not related to time limitations can be studied on sentences
where no timeouts occurred (NT).
Here the effects of the
extensions diverge: for the first linkage, performance with the UMLS
extension and the POS extension with the Brill tagger essentially
matches that of the unmodified LGP, while performance with xMG and
GENIA Tagger remains better. For the best linkage, we
observe a negative effect from the UMLS extension, indicating that for
some words the unknown word handling mechanism of LGP finds correct
links that are not allowed by the linking requirements given to those
words in the extended dictionary. This suggest that some errors
have occurred in the automatic mapping process\footnote{
An example of one such error is in the mapping of abbreviations (e.g. \emph{MHC}) 
to countable nouns, leading to failures to parse in the absence of
determiners.
}.
We
similarly observe the expected decrease in performance for the Brill
tagger for the best linkage, reflecting tagging errors. 

Even for the best linkage in sentences where no timeouts occurred,
the performance with the xMG extension and the POS extension with
GENIA Tagger is better than that of the original LGP. These extensions
can thus assign more appropriate linking requirements for some words
than the unknown word system of LGP. This indicates high tagging accuracy
for GENIA Tagger 
as well as an appropriate choice of linking requirements for
both extensions, and suggests some limitation in the unknown word
system of LGP.

Despite significant improvements in parsing performance, the best
performance achieved by any LGP extension is 88\%. This may again
suggest a limit on what performance can be achieved through the
lexical adaptation approaches.


\subsection{Combinations of the Extensions}
                       
 
The UMLS, xMG and POS tagging extensions are
to some extent complementary as their coverage of the corpus vocabulary does not completely overlap.
The dictionary extension provides the most frequent domain-specific lexicon while
the xMG extension has the advantage of being able to handle non-canonical 
 (e.g.\ \textit{mutation/deletion}, \textit{DNA-regions}) and rare words
 and misspellings. The POS extension can
benefit from the context-sensitiveness of the tagger to disambiguate words.

We evaluated all possible combinations of the three extensions.  In
these experiments we only used GENIA Tagger for the POS extension.
The results are given in tables~\ref{table:combination_performance}
and~\ref{table:combination_ambiguity}.

\begin{table}[h!t]
\centering
\begin{small}
\begin{tabular}{l|r|r|r|r|r}
                    &       & UMLS    & xMG     & UMLS               \\
Metric              & Orig  & \& xMG   & \& POS & \& POS   & All 3   \\ \hline
Time                & 15.4s & 9.5s    & 8.7s   & 8.3s     & 8.4s    \\
Lkg.\ ratio         & 1     & 0.67    & 0.59   & 0.62     & 0.66    \\
\end{tabular}
\caption{Ambiguity for combinations of the extensions.}
\label{table:combination_ambiguity}
\end{small}
\end{table}

On ambiguity, we observe small advantages for many of the
combinations, but rarely more than a 10\% reduction for either metric
compared to the simple extensions.
The effect of the combinations on overall performance is mixed. While all
combinations outperform the original LGP, combinations involving the 
UMLS extension appear to perform worse than those that do not,
while combinations involving the xMG and POS extensions perform
better. For sentences where no timeouts occurred the effect is simple:
for the best linkage, all combinations involving the UMLS
extension perform worse than the original LGP; only the combination
of the xMG and POS extensions is better.

The performance of the best combination approach essentially matches
that of the POS extension with GENIA Tagger alone, suggesting
that no further benefit can be derived from combinations
when an accurate domain tagger is available.


\section{Conclusions and Future Work}

We have studied three lexical adaptation approaches addressing 
biomedical domain vocabulary not
found in the lexicon of
the Link Grammar Parser:
automatic lexicon expansion, surface clue based morpho-guessing, and
the use of a POS tagger. 
We found that in a time-limited setting, any
approach resolving unknown words can improve efficiency and overall
performance.
In more detailed evaluation,
we found that the
automatic dictionary extension and the use of a general English POS
tagger can reduce 
performance, while the morpho-guessing approach
and the use of a domain-specific POS tagger had only positive effects.
We found no further benefit from combinations of the three
approaches.

Generally, our results suggest that when available, a high-quality
domain POS tagger is the best solution to unknown word issues in the
domain adaptation of a general parser, here providing an overall 10\%
relative reduction in error combined with a 45\% decrease in parsing
time. In the absence of such a resource, the use of a general
POS tagger is a poor substitute, and can lead to decreased
performance.  The use of heuristic methods for lexicon expansion 
carries the risk of mapping errors and should be accompanied by an
evaluation of the effect on parsing performance.  Conversely,
surface clues can provide remarkably good coverage and performance
when tuned to the domain, here using as few as 23 new rules.

Our implementation of the adaptations to LGP combines the
morpho-guessing extension with the capability to use information from
a POS tagger. Thus, the adapted parser is faster and more accurate
than the unmodified LGP in parsing biomedical texts both when used as
such and when used together with a domain POS tagger.  Further, both
extensions are implemented so that defining other morpho-guessing
rules and POS-mappings is straightforward, facilitating adaptation of
the modified parser also to other domains.  The adapted LGP is
available under an open-source licence at
\texttt{http://www.it.utu.fi/biolg}.

While we found that the considered approaches can significantly
improve efficiency and parsing performance, our results also indicate
some limitations for lexical adaptation. As future work, complementary
approaches addressing grammar adaptation, text preprocessing, handling
of complex terms, improved
parse ranking and named entity recognition can be considered to further
improve the applicability of LGP to the biomedical domain.

\section*{Acknowledgements}

The work of Sampo Pyysalo 
has been supported by Tekes,
the Finnish Funding Agency for Technology and Innovation.
\vspace{-0.2cm} 

\bibliographystyle{acl} \bibliography{llbibliography}

\end{document}